\def\year{2022}\relax
\documentclass[letterpaper]{article} 
\usepackage{aaai22}  
\usepackage{times}  
\usepackage{helvet}  
\usepackage{courier}  
\usepackage[hyphens]{url}  
\usepackage{graphicx} 
\usepackage{natbib}  
\usepackage{caption} 
\DeclareCaptionStyle{ruled}{labelfont=normalfont,labelsep=colon,strut=off} 
\usepackage{algorithm}
\usepackage{algorithmic}

\usepackage{newfloat}
\usepackage{listings}
\floatstyle{ruled}
\newfloat{listing}{tb}{lst}{}
\floatname{listing}{Listing}

\usepackage{amsmath,  amssymb, amsthm}
\usepackage{dsfont}
\usepackage{xcolor, tikz}
\usepackage{hyperref}
\usepackage{wrapfig}
\usepackage{stfloats}

   \newcommand{\reals}{\mathbb{R}}
   \newcommand{\naturals}{\mathbb{N}}
   \newcommand{\Ex}{\mathbb{E}}
   \renewcommand{\Pr}{\mathbb{P}}
   \newcommand{\Lo}[1]{{\mathcal L_{#1}}}
   \newcommand{\bLo}[1]{{\mathcal L^{0/1}_{#1}}}
   \newcommand{\sLo}[1]{{\mathcal L^{\to}_{#1}}}
    \newcommand{\scLo}[1]{{\mathcal L^{\to,\perp}_{#1}}}
    \newcommand{\sLoprime}[1]{{\mathcal L^{\leadsto}_{#1}}}
    \newcommand{\scLoprime}[1]{{\mathcal L^{\leadsto,\perp}_{#1}}}
    \newcommand{\gLo}[1]{{\mathcal L^{\mathrm{gr}}_{#1}}}
   \newcommand{\lo}{\ell}
   \newcommand{\blo}{{\ell^{0/1}}}
   \newcommand{\slo}{{\ell^{\to}}}
   \newcommand{\sloprime}{{\ell^{\leadsto}}}
   \newcommand{\sclo}{{\ell^{\to,\perp}}}
   \newcommand{\glo}{{\ell^{\mathrm{gr}}}}
   \newcommand{\scloprime}{{\ell^{\leadsto,\perp}}}

    \newcommand{\dHPX}{{d_{\mathcal{H},P_\X}}}

   \newcommand{\indct}[1]{\mathds{1}\left[{#1}\right]}
   
   \renewcommand{\P}{{\mathcal P}}
   \newcommand{\A}{{\mathcal A}}

  \newcommand{\F}{{\mathcal F}}
  \newcommand{\G}{{\mathcal G}}
  \renewcommand{\H}{{\mathcal H}}
  
  \newcommand{\U}{\mathcal{U}}
  \newcommand{\V}{\mathcal{V}}
  \newcommand{\X}{\mathcal{X}}
  \newcommand{\Y}{\mathcal{Y}}
  \newcommand{\Z}{\mathcal{Z}}
  \newcommand{\M}{\mathcal{M}}

  \newcommand{\Hcal}{{\mathcal H}}

  \newcommand{\marginal}{P_\X}
  \newcommand{\vc}{\mathrm{VC}}
  \newcommand{\opt}{\mathrm{opt}}
  \newcommand{\bopt}[1]{\mathrm{opt}^{0/1}_{#1}}

  \newcommand{\iid}{i.i.d.~}

  \newcommand{\cost}{\mathrm{cost}}
  \newcommand{\br}{\mathrm{br}}
  \newcommand{\burden}{\mathrm{brd}}
  \newcommand{\argmax}{\mathrm{argmax}}

  \newcommand{\x}{\mathbf{x}}
  \newcommand{\z}{\mathbf{z}}
  
\makeatletter
\newtheorem*{rep@theorem}{\rep@title}
\newcommand{\newreptheorem}[2]{
\newenvironment{rep#1}[1]{
 \def\rep@title{#2 \ref{##1}}
 \begin{rep@theorem}}
 {\end{rep@theorem}}}
\makeatother
  
\newtheorem{observation}{Observation}
\newreptheorem{observation}{Observation}
\newtheorem{theorem}{Theorem}
\newreptheorem{theorem}{Theorem}
\newtheorem{lemma}{Lemma}
\newreptheorem{lemma}{Lemma}
\newtheorem{definition}{Definition}

\newreptheorem{corollary}{Corollary}
\newtheorem{example}[theorem]{Example}

\newenvironment{proofof}[1]{\noindent{\bf Proof of {#1}:~~}}{\(\qed\)}
\newcommand{\BPFOF}{\begin{proofof}} \newcommand {\EPFOF}{\end{proofof}}

\nocopyright

\title{Learning Losses for Strategic Classification}

\author {
    Tosca Lechner,\textsuperscript{\rm 1}
    Ruth Urner, \textsuperscript{\rm 2}
}
\affiliations {
    \textsuperscript{\rm 1} University of Waterloo, Cheriton School of Computer Science, Waterloo, Canada, tlechner@uwaterloo.ca\\
    \textsuperscript{\rm 2} York University, Lassonde School of Engineering, EECS Department, Toronto, Canada, ruth@eecs.yorku.ca\\
   
}

\usepackage{bibentry}

\begin{document}

\maketitle

\begin{abstract}
Strategic classification, i.e. classification under possible strategic manipulations of features, has received a lot of attention from both the machine learning and the game theory community. Most works focus on analysing properties of the optimal decision rule under such manipulations. In our work we take a learning theoretic perspective, focusing on the sample complexity needed to learn a good decision rule which is robust to strategic manipulation. We perform this analysis by introducing a novel loss function, the \emph{strategic manipulation loss}, which takes into account both the accuracy of the final decision rule and its vulnerability to manipulation. We analyse the sample complexity for a known graph of possible manipulations in terms of the complexity of the function class and the manipulation graph. Additionally, we initialize the study of learning under unknown manipulation capabilities of the involved agents. Using techniques from transfer learning theory, we define a similarity measure for manipulation graphs and show that learning outcomes are robust with respect to small changes in the manipulation graph. Lastly, we analyse the (sample complexity of) learning of the manipulation capability of agents with respect to this similarity measure, providing novel guarantees for strategic classification with respect to an unknown manipulation graph.
\end{abstract}

\section{Introduction}
In many scenarios where a decision rule is learned from data, the publication of this decision rule has an effect on the distribution of the underlying population that may harm the quality of the rule. 
For example, applicants for a loan may change details in their bank account to receive a better score, people may join a gym or sports club without ever intending to participate, in order to get a better health insurance policy, or students may employ different strategies such as registering to volunteer, or joining rare clubs (without attending either) to appear better on college applications.

Effects and incentives resulting from strategic behavior in classification scenarios have received substantial attention from both machine learning and game-theoretic perspectives in recent years \cite{HardtMPW16,MilliMDH19,HaghtalabILW20,TsirtsisR20,ZhangC21}. Most works study this as a two-player game between an institution that publishes a decision rule and a  population of best responding agents to be classified. Given the classifier, these agents may change their feature representations in order to obtain a more favorable classification outcome. To prevent the induced additional classification error the institution will publish a modified predictor, not transparently reflecting the underlying intent and potentially causing additional harm to sub-populations that may be less equipped to perform the required changes to their representations \cite{HuIV19}.

In this work, we propose a learning theoretic take on this scenario. In machine learning, it is common to model desiderata for a learning outcome in form of a loss function. The goal of the learning process is then to identify a predictor that minimizes this loss in expectation over a data-generating distribution. 
Thus, we here define a novel loss function for learning under strategic manipulations.
The aim of this loss is to induce a combination of two (potentially competing) requirements: achieving low classification error taking into account that individuals being classified may manipulate their features, and discouraging such feature manipulations overall. Prior work has shown that these may be conflicting requirements \cite{ZhangC21}. Our proposed loss function thus aims to induce a balanced combination of these requirements rather than strictly enforcing one, and then only observing the effect on the other (as is implied by frameworks that aim to minimize classification error under best-responding agents \cite{HardtMPW16,MilliMDH19} or enforcing incentive-compatibility \cite{ZhangC21}).

To define our \emph{strategic manipulation loss} we employ an abstraction of the plausible feature manipulations in form of a \emph{manipulation graph} \cite{ZhangC21}. An edge $\x\to\x'$ in this graph indicates that an individual with feature vector $\x$ may change their features to present as $\x'$ if this leads to a positive classification, for example since the utility of this change in classification exceeds the cost of the change between these vectors. We define our strategic loss in dependence of this graph and carefully motivate the proposed loss in terms of requirements and effects from previous literature. We then analyze the sample complexity of learning with this loss function. We identify sufficient conditions for proper learnability that take into account the interplay between a hypothesis class and an underlying manipulation graph. Moreover, we show that every class that has finite VC-dimension is learnable with respect to this loss by drawing a connection to results in the context of learning under adversarial perturbations \cite{MontasserHS19}. This effect may be surprising, since it presents a contrast to learning VC-classes with the sole requirement of minimizing classification error under strategic feature manipulations, which has been shown can lead to some VC-classes not being learnable \cite{ZhangC21}. Thus, our analysis shows that balancing classification error with disincentivizing feature manipulations can reduce the complexity of the learning problem.

Moreover, we show that the quality of learning outcomes under our loss function is robust to inaccuracies in the manipulation graph. Such a robustness property is important, since an assumed graph might not exactly reflect agents' responses. In fact, it has recently been argued that the model of best-responding agents is not backed up by empirical observations on agent distributions after strategic responses \cite{JagadeesanMH21}. Moreover, different sub-populations may have differences in their manipulation graphs (different capabilities to manipulate their features) or a manipulation graph may be inferred from data and therefore exhibit statistical errors. We introduce a novel distance measure between manipulation graphs by drawing connections to learning bounds in transfer learning \cite{bendavid2010differentdomains,mansour2009domainadaptation} and show that the strategic loss of a learned predictor when employing a different manipulation graph can be bounded in terms of this distance measure. 
Finally, we present some initial results on how manipulation graphs may be learned from data.
\subsection{Related work}
That learning outcomes might be compromised by agents responding to published classification rules with strategic manipulations of their feature vectors was first pointed out over a decade ago \cite{DalviDMSV04,BrucknerS11} and has received substantial interest from the research community in recent years initiated by a study by Hardt et al. that differentiated the field from the more general context of learning under adversarial perturbations \cite{HardtMPW16}. That study considered strategic responses being induced by separable cost functions for utility maximizing agents and studied the resulting decision boundaries for certain classes of classifiers. Recent years have seen a lot of interest in better understanding the interplay of various incentives in settings where a decision rule is published and thereby has an effect on how the entities that are to be classified might present themselves to the decision-maker. In particular, various externalities to this scenario have been analyzed. A general cost to society formalized in form of a ``social burden'' incurred by the costs of enforced feature manipulation, has been shown to occur when institutions anticipate strategic responses \cite{MilliMDH19,JagadeesanMH21}. Further, it has been demonstrated how such a burden may be suffered to differing degrees by various subgroups of a population that may differ in their capabilities to adapt their features in ways that are favorable to them \cite{MilliMDH19,HuIV19}, raising concerns over fairness in such scenarios.

Recent studies have extended the original game-theoretic model of a classifier publishing intuition and best responding subjects. For example, a recent work studied how strategic modification can also be a positive effect and how that should be taken into consideration by the institution \cite{HaghtalabILW20}. Such a perspective has been connected to underlying causal relations between features and classification outcome and resulting strategic recommendations \cite{MillerMH20,TsirtsisR20}. Further, a very recent study has explored how the model of a best responding agent may be relaxed to better reflect empirically observed phenomena \cite{JagadeesanMH21}.

Much of previous work considers the scenario of classification with strategic agents on a population level. A few recent studies have also analyzed how phenomena observed on samples reflect the underlying population events \cite{HaghtalabILW20}. Notably, very recent studies provided first analyses of learning with strategically responding agents in a PAC framework \cite{ZhangC21, SundaramVXY21}. The former work studied the sample complexity of learning VC-classes in this setup and analyzed effects on sample complexity of enforcing incentive compatibility for the learned classification rules. Our work can be viewed as an extension of this analysis. We propose to combine aspects of incentive compatibility and minimizing negative externalities such as social burden in form of a novel loss function that may serve as a learning objective when strategic responses are to be expected.

 Our sample complexity analysis is then hinging on techniques developed in the context of learning under adversarial perturbations,
a learning scenario which has received considerable research attention in recent years \cite{feige2015learning,cullina2018pac,MontasserHS19,MontasserHS21}. While the learning problems are not identical, we present how strategic behaviour can be modeled as a form of ``one-sided adversarial perturbation'' and inheritance of resulting learning guarantees.

\subsection{Overview on contributions}
In Section~\ref{sec:setup} we review our notation and then introduce our new notion of strategic loss and motivate it. 
Our main contributions can be summarized as follows:
\begin{description}
\item[Strategic manipulation loss] We propose a novel loss function for learning in the presence of strategic feature manipulations. We carefully motivate this loss by relating it to concepts of social burden and incentive compatibility (and their potential trade-offs with accuracy) in prior literature.
\item[Sample complexity analysis] We analyze (PAC type) learnability of VC-classes with the strategic loss. We provide sufficient conditions (and examples of when they are satisfied) for learnability with a proper learner. By drawing connections and adapting results from learning under adversarial perturbations to our setup, we also show that, while proper learnability can not always be guaranteed, every VC-class is learnable under the strategic loss with an improper learner.
\item[Robustness to inaccurate manipulation information] We investigate the impact of using an approximate manipulation graph to yield a surrogate strategic loss function for cases where the true manipulation graph is not known or not accessible. For this, we introduce a novel similarity measure on graphs and show that if graphs are similar with respect to our notion then they yield reasonable surrogate strategic losses for each other (Theorem~\ref{thm:approximategraphbound}).
\item[Learning the manipulation graph]  We explore the question of whether it is possible to learn a manipulation graph that yields a good surrogate strategic loss. We identify a sufficient condition for a class of graphs $\G$ being learnable with respect to our previously defined similarity measure for graphs (Theorem~\ref{thm:dHPXlearning}), which in turn guaranteed the learning of a reasonable surrogate loss.
\end{description}
All proofs can be found in the appendix of the full version.

\section{Setup}\label{sec:setup}

\subsection{Basic Learning Theoretic Notions for Classification}
We employ a standard setup of statistical learning theory for classification.
We let  $\X\subseteq\reals^d$ denote the domain and $\Y$ (mostly $\Y=\{0,1\}$) a (binary) label space.
We model the data generating process as a distribution $P$ over $\X\times \Y$ and let $P_\X$ denote the marginal of $P$ over $\X$.
We use the notation $(\x,y)\sim P$ to indicate that $(\x,y)$ is a sample from distribution $P$ and $S\sim P^n$ to indicate that set $S$ is a sequence (for example a training or test data set) of $n$ \iid samples from $P$.
Further, we use notation $\eta_P(\x) = \Pr_{(\x,y)\sim P}[y = 1 \mid \x]$ to denote the \emph{regression} or \emph{conditional labeling function} of $P$.
We say that the distribution has \emph{deterministic labels} if $\eta_P(\x) \in \{0,1\}$ for all $\x\in\X$.

A \emph{classifier} or \emph{hypothesis} is a function $h:\X\to\Y$.
A classifier $h$ can naturally be viewed a subset of $\X\times \Y$, namely $h = \{(\x,y)\in \X\times \Y ~\mid~ \x\in \X,~y = h(\x)\}$.
We let $\F$ denote the set of all Borel measurable functions from $\X$ to $\Y$ (or all functions in case of a countable domain). 
A \emph{hypothesis class} is a subset of $\F$, often denoted by $\H\subseteq \F$.
For a loss function $\lo: \F \times \X\times \Y \to \reals$ we denote the expected loss for a distribution $P$ as $\Lo{P}$ and the empirical loss for a sample $S$ as $\Lo{S}$. We use standard definitions like PAC learnability, sample complexity, and approximation error. For further elaborations on these definitions, we refer the reader to the appendix for an extended definitions section or a textbook \cite{shalev2014understanding}.

\subsection{Strategic Classification}\label{ss:strategiclossdefinition}
\subsubsection{Learning objectives in prior work} The possibilities for strategic manipulations of a feature vector are often modeled in terms of a cost function $\cost: \X \times \X \to \reals_0^+$, so that $\cost(\x, \x')$ indicates how expensive it is for an individual with feature vector $\x$ to present as $\x'$. A natural minimal assumption a cost function should satisfy is $\cost(\x, \x) = 0$ for all feature vectors $\x$. It is then typically assumed that instances best-respond to a published classifier, in that the individual $\x$ would choose to pay the cost of presenting as $\x'$ as long as the cost doesn't exceed the utility that would be gained from the difference in classification outcome. Assuming the benefit of individual $\x$ receiving classification $1$ over classification $0$ is $\gamma$, the manipulation would happen if $\cost(\x, \x') \leq \gamma$ and $h(\x) = 0$ while $h(\x') = 1$ for a given classifier. That is, we can define the best response of an individual with feature vector $\x$ facing classifier $h$ as
\[
\br(\x, h) = \argmax_{\x'\in\X} [\gamma\cdot h(\x') - \cost(\x, \x')],
\]
with ties broken arbitrarily, and assuming that, if the original feature vector $\x$ is among those maximizing the above, then the individual would choose to maintain the original features. An often assumed learning goal is then \emph{performative optimality} \cite{PerdomoZMH20,JagadeesanMH21}, which stipulates that a learner should aim to maximize accuracy on the distribution it induces via the agent responses. That is, this objective can be phrased as minimizing
\[
\Ex_{(\x, y) \sim P}\indct{h(\br(\x, h)) \neq y}
\]
An alternative view on this setup, if the agent responses are deterministic, is to  view the above as minimizing the binary loss of the \emph{effective hypothesis} $\hat{h}:\X \to \{0,1\}$ that is induced by $h$ and the agents' best responses $\br(\cdot, \cdot)$ \cite{ZhangC21}, defined as

\begin{equation}\label{eq:effectivehypothesis}
\hat{h}(\x) = h(\br(\x, h)).
\end{equation}

The goal of performative optimality has been combined with the notion of \emph{social burden} that is induced by a classifier \cite{MilliMDH19,JagadeesanMH21}. This notion reflects that it is undesirable for a (truly) positive instance to be forced to manipulate its features to obtain a (rightfully) positive classification. This is modeled by considering the \emph{burden} on a positive individual to be the cost that is incurred by reaching for a positive classification and the \emph{social burden} incurred by a classifier to be the expectation with respect to the data-generating process over these costs:
\[
\burden_P(h) = \Ex_{(\x, y)\sim P}\left[\min_{\x'\in\X}\{\cost(\x, \x') \mid h(\x') = 1\} ~\mid~ y = 1 \right]
\]
It has been shown that optimizing for performative optimality (under the assumption of deterministic best-responses) also incurs maximal social burden \cite{JagadeesanMH21}.

\subsubsection{A new loss function for learning under strategic manipulations} 
Arguably, to seek performative optimality (or minimize the binary loss over the effective hypothesis class) the cost function as well as the value $\gamma$ (or function $\gamma:\X \to \reals$) of positive classification needs to be known (or at least approximately known). To take best responses into account, a learner needs to know what these best responses may look like. In that case, we may ignore the details of the cost function and value $\gamma$, and simply represent the collection of \emph{plausible manipulations} as a directed graph structure $\M = (\X, E)$ over the feature space $\X$ \cite{ZhangC21}. The edge-set $E$ consists of all pairs $(\x, \x')$ with $\cost(\x,\x') \leq \gamma$, and we will also use the notation $\x\to \x'$ for $(\x, \x')\in E$, and write $\M = (\X, E) = (\X, \to)$. We note that this formalism is valid for both countable (discrete) and uncountable domains.

Given the information in the so obtained \emph{manipulation graph $\M = (\X, \to)$}, we now design a loss function for classification in the presence of strategic manipulation that reflects both classification errors and the goal of disincentivizing manipulated features as much as possible.
Our proposed loss function below models that it is undesirable for a classifier to assign $h(\x) = 0$ and $h(\x')=1$ if feature vector $\x$ can present as $\x'$. This is independent of a true label $y$ (e.g. if $(\x, y)$ is sampled from the data generating process). If the label $y = 0$ is not positive, the point gets misclassified when $\x$ presents as $\x'$. On the other hand, if the true label is $1$, then either a true positive instance is forced to manipulate their features to obtain a rightly positive outcome (and this contributes to the social burden), or, if the choice is to not manipulate the features, the instance will be misclassified (prior work has also considered models where true positive instance are ``honest'' and will not manipulate their features \cite{DongRSWW18}). 
Here, we propose to incorporate both misclassification and contributions to social burden into a single loss function that a learner may aim to minimize.

\begin{definition}\label{def:strategicloss}
We define the \emph{strategic loss} $\slo: \F \times \X \times \Y \to \{0,1\}$ with respect to manipulation graph $(\X, \to)$ as follows:
\[
\slo(h, \x, y) = \left\{\begin{array}{ll}
    1 & \text{if } h(\x) \neq y\\
    1 & \text{if } h(\x) = 0 \text{ and }\\
    &\exists \x' \text{ with } \x\to \x' \text{ and } h(\x') =1\\
    0 & \text{else} \\
\end{array}\right.
\]
\end{definition}
Note that the first two cases are not mutually exclusive. 
The above loss function discretizes the social burden by assigning a loss of $1$ whenever a positive individual is required to manipulate features. 
As for the standard classification loss, the above point-wise definition of a loss function allows to define the \emph{true strategic loss} $\sLo{P}(h)$ and \emph{empirical strategic loss} $\sLo{S}(h)$ of a classifier with respect to a distribution $P$ or a data sample $S$. 

\subsection{Comparison with Alternative Formalisms for Strategic Classification}\label{ss:settingcomparison}

To motivate our proposed loss, we here discuss several scenarios where, we'd argue, minimizing the strategic loss leads to a more desirable learning outcome than learning with a binary loss, while taking strategic manipulations into account.  
As discussed above, a common approach to modeling classification in a setting where strategic manipulations may occur is to assume that all agents will best-respond to a published classifier. That is, if $h(\x) = 0$,  $h(\x') = 1$ and $\x\to \x'$, then the agent with initial feature vector $\x$ will \emph{effectively} receive classification $1$. 
A natural modeling is then to consider the effective hypothesis $\hat{h}$ induced $h$ (see Equation \ref{eq:effectivehypothesis}) and aim to minimize the classification error with the \emph{effective class} $\hat{\H} = \{ f \mid f = \hat{h} \text{ for some } h\in \H\}$ \cite{ZhangC21}. However it has been shown, that the VC-dimension of $\hat{\H}$ may be arbitrarily larger than the VC-dimension of $\H$, and may even become infinite \cite{ZhangC21}. When learning this effective class $\hat{\H}$ with respect to the binary loss (which corresponds to aiming for performative optimality), this will imply that the class is not learnable. By contrast, we will show below that any class of finite VC-dimension remains learnable with respect to the strategic loss (Theorem \ref{thm:adaptationofmontasser}). 

It has also been shown that the negative effects in terms of sample complexity of considering the effective hypothesis class can be avoided by considering only \emph{incentive-compatible} hypotheses in $\H$, that is outputting only such hypotheses that will not induce any feature manipulations in response to the published classifier \cite{ZhangC21}. While this avoids the growths in terms of VC-dimension, it may prohibitively increase the approximation error of the resulting (pruned) class as we show in the example below. We would argue that this illustrates how low sample complexity, in itself, is not a sufficient criterion for learning success.

\begin{example}
Consider $\X = \naturals$ and a manipulation graph that includes edges $n\to n+1$ and $n \to n-1$ for all $n\in \naturals$. This is a reasonable structure, considering that the cost of moving the (one-dimensional) feature by $1$ is worth a positive classification outcome. However, the only two hypotheses that are incentive compatible in this case are the two constant functions $h_0:\X\to\{0\}$ and $h_1:\X\to\{1\}$. Thus, requiring incentive compatiblity forces the learner to assign all points in the space with the same label. This class, in fact, has low sample complexity. However, arguably, restricting the learning to such a degree (and suffering the resulting classification error, which will be close to $0.5$ for distributions with balanced classes), is, in most cases not a reasonable price to pay for dis-incentivising feature manipulations. 
\end{example}

The following example illustrates how our loss function can be viewed as incorporating the notion of social burden directly into the loss. 

\begin{example}
Let's again consider a domain $\X = \naturals$ and a manipulation graph $\M$ with edges $n\to n+1$ for all $n\in\naturals$. We consider distributions that have support $\{(1, 0), (2,0), (3,1), (4,1)\}$, thus only these four points have positive probability mass and a hypothesis class of thresholds $\H = \{h_a \mid a\in\reals\}$, with $h_a(\x) = \indct{\x \geq a}$. The true labeling on these distributions is $\eta(\x) = h_{2.5}(\x)$. On all distributions, where all four points have positive mass the performatively optimal hypothesis (or effective hypothesis of minimal binary loss) however is $h_{3.5}$. The social burden incurred then is $\burden_P(h_{3.5}) = P((3,1))\cdot\cost(3,4)$. It is important to note that the performative optimality of $h_{3.5}$ is independent of the distribution $P$ over the points. A learner that minimizes the strategic loss, on the other hand, will take the distribution $P$ into account and output $h_{2.5}$ if $P((2,0)) < P((3,1))$, while outputting $h_{3.5}$ if $P((2,0)) > P((3,1))$. If the difference in mass of these points (or the margin areas in a more general setting) is significant, then minimizing the strategic loss will opt for allowing a small amount of manipulation in turn for outputting a correct classification rule in case $P((2,0)) \ll P((3,1))$; and it will opt for changing the classification rule, accept a small amount of social burden in exchange for preventing a large amount of manipulations and resulting classification errors, in case $P((2,0)) \gg P((3,1))$. We would argue that this reflects a desirable learning outcome.
\end{example}

\section{Learnability with the Strategic Loss}\label{s:learnabilityanalysis}
\subsection{Warm up: Loss Classes and Learnability}\label{ss:lossclasses}
It is well known that a class $\H$ is learnable (with respect to the set of all distributions) if the \emph{loss class} induced by a $0/1$-valued loss function $\lo$ has finite VC-dimension. In the case of the binary classification loss, this is in fact a characterization for learnability (and the VC-dimension of the loss class is identical to the VC-dimension of the hypothesis class $\H$). In general, bounded VC-dimension of the loss class is a sufficient condition for learnability (the VC-dimension provides an upper bound on the sample complexity), but it is not a necessary condition (it doesn't, in general, yield a lower bound on the sample complexity of learning a class $\H$ with respect to some loss $\lo$). We start by reviewing these notions for the classification loss and then take a closer look at the loss class induced by the strategic loss.

Let $\lo$ be a loss function and $h$ be a classifier. We define the \emph{loss set} $h_\lo \subseteq \X\times \Y$ as the set of all labeled instances $(\x,y)$ on which $h$ suffers loss $1$.
The \emph{loss class} $\H_\lo$ is the collection of all loss sets (in the literature, the loss class is often described as the function class of indicator functions over these sets). In the case of binary classification loss $\blo$, the loss set of a classifier $h$ is exactly the complement of $h$ in $\X\times \Y$. That is, in this case the loss set of $h$ is also a binary function over the domain $\X$ (namely the function $\x\mapsto |h(\x) -1|$). For the strategic loss on the other hand, the loss set of a classifier $h$ is not a function, since it can contain both $(\x,0)$ and $(\x,1)$ for some points $\x\in\X$, namely if $h(\x) =0$ and there exists an $\x'$ with $\x\to\x'$ and $h(\x') =1$. For a class $\H$ we let $\H_\blo$ denote the loss class with respect to the binary loss and $\H_{\slo}$ the loss class with respect to the strategic loss.

\begin{definition}\label{def:vc}
Let $\Z$ be some set and $\U\subseteq 2^\Z$ be a collection of subsets of $\Z$. 
We say that a set $S\subseteq \Z$ is \emph{shattered by $\U$} if
\[
\{U \cap S \mid U\in \U\} = 2^S,
\]
that is, every subset of $S$ can be obtained by intersecting $S$ with some set $U$ from the collection $\U$. 
The \emph{VC-dimension} of $\U$ is the largest size of a set that is shattered by $\U$ (or $\infty$ if $\U$ can shatter arbitrarily large sets).
\end{definition}

It is easy to verify that for the binary loss, the VC-dimension of $\H$ as a collection of subsets of $\X\times\Y$ is identical with the VC-dimension of $\H_\blo$ (and this also coincides with the VC-dimension of $\H$ as a binary function class \cite{shalev2014understanding}; VC-dimension is often defined for binary functions rather than for collection of subsets, however, this is limiting for cases where the loss class is not a class of functions).

We now show that the VC-dimension of a class $\H$ and its loss class with respect to the strategic loss can have an arbitrarily large difference. Similar results have been shown for the binary loss class of the effective class $\hat{\H}$ induced by a manipulation graph \cite{ZhangC21}. However the binary loss class of $\hat{\H}$ is different from the strategic loss class of $\H$ and, as we will see, the implications for learnability are also different. 

\begin{observation}\label{obs:vcgrows}
For any $d\in\naturals\cup\{\infty\}$ there exists a class $\H$ and a manipulation graph $\M = (\X, \to)$ with $\vc(\H) =1 $ and $\vc(\H_\slo) \geq d$.
\end{observation}

On the other hand, we prove that the VC-dimension of the strategic loss class $\H_{\slo}$ is always at least as large as the VC-dimension of the original class.

\begin{observation}\label{obs:VCinequ}
For any hypothesis class $\H$ and any manipulation graph $\M = (\X, \to)$, we have $\vc(\H) \leq \vc(\H_\slo)$.
\end{observation}
 
Standard VC-theory tells us that, for the binary classification loss, any learner that acts according to the  ERM (Empirical Risk Minimization) principle is a successful learner for classes of bounded VC-dimension $d$.
For a brief recap of the underpinnings of this result, we refer the reader to the supplementary material or for further details to \cite{shalev2014understanding}. 
In the case of general loss classes with values in $\{0,1\}$, the VC-dimension does not characterize learnability.
In particular, we next show that the VC-dimension of the strategic loss class does not imply a lower bound on the sample complexity.

\begin{theorem}\label{thm:nolowerbound}
For every $d\in\naturals\cup\{\infty\}$, there exists a hypothesis class $\H$ with $\vc(\H_\slo) = d$ that is learnable with sample complexity $O(\log(1/\delta)/\epsilon)$ in the realizable case.
\end{theorem}

\subsection{Sufficient Conditions for Strategic Loss Learnability}

In the previous section, we have seen that the loss class having a finite VC-dimension is a sufficient (but not necessary) condition for learnability with respect to the strategic loss. We have also seen that the VC-dimension of $\H_{\slo}$ can be arbitrarily larger than the VC-dimension of $\H$.
To start exploring what determines learnability under the strategic loss, we provide a sufficient condition for a class to be properly learnable with respect to the strategic loss.

Note that for a hypothesis $h$, the strategic loss set $h_\slo$ can be decomposed into the loss set of $h$ with respect to the binary loss and the component that comes from the strategic manipulations. Formally, we can define the \emph{strategic component loss}.
\begin{definition}\label{def:sclo}
We let the strategic component loss with respect to manipulation graph $\to$ be defined as
\[\sclo(h,\x) = \indct{h(\x)= 0 \land \exists \x': \x\to \x',\ h(\x') = 1}\]
We note that $\slo(h,\x,y) \leq \blo(h,\x,y) + \sclo(h,\x) $.
We will denote the true strategic component loss with respect to marginal distribution $P_\X$ as $\scLo{P_\X}$.

\end{definition}

For the loss sets, we then get
\[
h_{\blo} = \{(\x, y) \in \X\times\Y ~\mid~ h(\x) \neq y \},
\]
and
\begin{align*}
h_{\sclo} = \{(\x, y) \in \X\times\Y ~\mid~ & h(\x) = 0 \\
& \land \exists \x': \x\to\x',\ h(\x') = 1 \}.    
\end{align*}
This implies 
\[
h_{\slo} = h_{\blo} \cup h_{\sclo}
\]
for all classifiers $h\in\F$, and thereby
\[
\H_{\slo} = \{ h_{\blo} \cup h_{\sclo} ~\mid~  h\in \H\}.
\]
By standard counting arguments on the VC-dimension of such unions (see, for example Chapter 6 of \cite{shalev2014understanding} and exercises in that chapter), it can be shown this decomposition implies that  $\vc(\H_{\slo}) \leq d\log d$ for $d = \vc(\H_{\blo}) + \vc(\H_{\sclo}) =  \vc(\H) + \vc(\H_{\sclo})$. Thus, if both the class $\H$ itself and the class of strategic components have finite VC-dimension, then $\H$ is properly learnable by any learner that is an ERM for the strategic loss:

\begin{theorem}\label{thm:boundedvc}
Let $\H$ be a hypothesis class with finite $\vc(\H) + \vc(\H_{\sclo}) = d < \infty$. Then $\H$ is properly PAC learnable with respect to the strategic loss (both in the realizable and the agnostic case).
\end{theorem}

Whether the class of strategic components has finite VC-dimension intrinsically depends on the interplay between the hypothesis class $\H$ and the graph structure of the manipulation graph. 
In Observation \ref{obs:vcgrows}, we have seen that the graph structure can yield the strategic component sets to have much larger complexity than the original class. In the appendix, Section B,
we provide a few natural examples, where the VC-dimension of the strategic components is finite.

Theorem \ref{thm:boundedvc} provides a strong sufficient condition under which any empirical risk minimizer for the strategic loss is a successful agnostic learner for a class of finite VC-dimension. We believe, in many natural situations the conditions in that theorem will hold, and analyzing in more detail which graph structure, combinations of graphs structures and hypothesis classes or classes of cost function lead to the strategic component sets having finite VC-dimension is an intriguing direction for further research.

We close this section with two results, both stated in Theorem \ref{thm:adaptationofmontasser}, on the learnability under the strategic loss in the general case where the VC-dimension of the strategic component sets may be infinite. First, there are classes and manipulation graphs for which no proper learner is (PAC-) successful, even in the realizable case. Second, for any class of finite VC-dimension and any manipulation graph, there exists an improper PAC learner. These results follow by drawing a connection from learning under the strategic loss to learning under an adversarial loss \cite{MontasserHS19}. In the general adversarial loss setup, every domain instance $\x$ is assigned a set of potential perturbation $\U(\x)$, and the adversarial loss of a hypothesis $h$ is then defined as
\[
\lo^{\U}(h, \x, y) = \indct{\exists \x'\in \U(\x): h(\x') \neq y}.
\]
The strategic loss can be viewed as a one-sided version of the adversarial loss, where the perturbation sets differ conditional on the label of a point, and where $\U(\x, 1) = \{\x\}$, while $\U(\x, 0) = \{\x' \in \X \mid \x \to \x'\}$. The following results on learnability with the strategic loss then follow by slight modifications of the corresponding proofs for learning under adversarial loss.

\begin{theorem}\label{thm:adaptationofmontasser}{\bf (Adaptation of Theorem 1 and Theorem 4 in \cite{MontasserHS19})}\\ 
There exists a hypothesis class $\H$ with $\vc(\H) = 1$ that is not learnable with respect to the strategic loss by any proper learner $\A$ for $\H$ even in the realizable case. On the other hand, every class $\H$ of finite  VC-dimension is learnable (by some improper learner).
\end{theorem}

\section{Strategic loss with respect to an approximate manipulation graph}\label{sec:approximategraph}
In many situations one might not have direct access to the true manipulation graph $\mathcal{M} = (V,E)$, but only to some approximate graph $\mathcal{M}' = (V, E')$. In this section, we will investigate how this change of manipulation graph impacts the corresponding loss function. We define a criterion for measuring the similarity of graphs with respect to hypothesis class $\H$ and show that similar graphs will yield similar strategic losses. That is, we show an upper bound on the true strategic loss of a hypothesis $h$ (i.e., strategic loss with respect to the true manipulation graph) in terms of the graph similarity and the surrogate strategic loss of $h$ (i.e., the strategic loss with respect to the approximate graph). We will use $\x \leadsto \x'$ to denote $(\x,\x')\in E'$. As the set of vertices $V$ is always equal to $\X$ in our setting, the graphs $\M$ and $\M'$ are uniquely defined by $\to$ and $\leadsto$ respectively. We will therefore use $\to$ and $\M$, as well as $\leadsto$ and $\M'$ interchangeably.

We now define the distance between graphs with respect to a hypothesis class $\H$ by the impact a change of manipulation graph has on the strategic component loss of elements of $\H$. 
This definition and its later use are inspired by works on domain adaptation \citep{bendavid2010differentdomains,mansour2009domainadaptation}.
\begin{definition}
For two manipulation graphs, given by $\to$ and $\leadsto$, we let their $\mathcal{H}$-$P_\X$-distance be defined as 
\[d_{\mathcal{H},P_\X}(\to, \leadsto) = \sup_{h\in \mathcal{H}} \mathbb{E}_{\x\sim P_\X}[|\sclo(h,\x) - \scloprime(h,\x) |]\]
\end{definition}

We will now bound the strategic manipulation loss $\sLo{P}(h)$ with respect to the true graph $\to$ in terms of the strategic manipulation loss $\sLoprime{P}(h)$ with respect to the approximate graph $\leadsto$ and the $\H$-$P_\X$-distance between $\to$ and $\leadsto$.

\begin{theorem}\label{thm:approximategraphbound}
Let $\mathcal{H}$ be any hypothesis class and $\to,\leadsto$ two manipulation graphs. Then for any distribution $P$ over $\X\times\Y$ and any $h\in \mathcal{H}$ we have
\begin{align*}
   \sLo{P}(h)  &\leq \bLo{P}(h) + \scLoprime{P}(h) + \dHPX(\to, \leadsto) \\
   &\leq  2 \sLoprime{P}(h) + \dHPX(\to, \leadsto) . 
\end{align*}
Furthermore, by rearranging the result, we get
\[ \frac{1}{2} \sLoprime{P}(h) - \dHPX(\to, \leadsto) \leq \sLo{P}(h). \]
\end{theorem}

We note that the expression $\dHPX(\to, \leadsto)$ is independent of the labeling and can therefore be estimated using data without any label information. Furthermore, we have seen that small $\dHPX(\to, \leadsto)$ tightens the upper as well as the lower bound on $\sLo{P}(h)$. Therefore, $\dHPX(\to, \leadsto)$ is a suitable distance measure for approximating the structure of the manipulation graph. In the following section, we will explore learning $\leadsto$ with low $\dHPX(\to, \leadsto)$ from finite samples. 
\section{Learning a manipulation graph}\label{sec:graphlearning}
In the last section, we have assumed to be given an approximate manipulation graph which we can use to learn a classifier with low strategic loss. We now want to go one step further and pose the goal of learning a manipulation graph $\leadsto$ from a predefined class of graphs $\G$ such that $\scloprime$ serves as a good strategic surrogate loss for $\sclo$. From Theorem~\ref{thm:approximategraphbound} we already know that $\scloprime$ is a good surrogate loss for $\sclo$ if $\dHPX(\to,\leadsto)$ is small. This section will thus focus on learning an approximate manipulation graph $\leadsto \in \G$ with small $\dHPX(\to,\leadsto)$.

In order to further specify our learning problem, we will now describe what the input of such a learning procedure will look like.
For a manipulation graph $\to$, let $B_{\to}: \X \rightarrow 2^{\X}$ be the function that maps an instance $\x$ to its set of children, i.e., $B_{\to}(\x) = \{\x'\in \X: \x\to \x'\} $. We note that a manipulation graph $\to$ is uniquely defined by $B_{\to}$. Thus we will sometimes use $B_{\to}$ and $\to$ interchangeably. The input to our learning procedure will be of the form of samples $S= \{(\x_1,B_{\to}(\x_1)), \dots , (\x_n,B_{\to}(\x_n))\}$ from the true manipulation graph $\to$.

As a next step in formulating our learning problem, we will need to define a loss function. As stated above, our goal is to learn $\leadsto \in \G$ with small $\dHPX(\to,\leadsto)$. As the definition of $\dHPX(\to,\leadsto)$ contains a supremum over all $h\in \H$, we cannot use it as a loss directly (as a loss needs to be defined point-wise). However, we can formulate a loss that is closely related and will serve to guarantee low $\dHPX(\to,\leadsto)$.
Let the \emph{graph loss} for a manipulation graph $\leadsto$, a domain point $x$, a manipulation set $B\subset \X$ and a hypothesis $h$ as
\[\glo(h,\leadsto,\x,B) = \begin{cases} 1 &\text{ if } h(\x)= 0 \land \exists \x'\in B: h(\x') =1 \\
&\land\forall \x'': \x\leadsto \x'' \text{ implies } h(\x'') =0\\
1 &\text{ if } h(\x)= 0 \land \forall \x'\in B: h(\x') =0 \\
&\land\exists \x'': \x\leadsto \x'' \text{ and } h(\x) =1 \\
0 & \text{ otherwise.} 
\end{cases} \]
This loss is indeed closely related to the $\H$-$P_\X$-distance as $ \glo(h,\leadsto,\x,B_{\to}(\x)) = |\sclo(h,\x) - \scloprime(h,\x)  | $.

The \emph{true graph loss} with respect to some marginal $P_\X$ and true manipulation graph $\to$ is then defined by
\[\gLo{(P_\X,\to) }(h,\leadsto) = \mathbb{E}_{\x\sim P_\X}[\glo(h,\leadsto,\x,B_{\to}(\x))].\]
Furthermore for a sample $S= \{(\x_1,B_1),\dots (\x_n,B_n)\}$ we define the \emph{empirical graph loss} as
\[\gLo{S }(h,\leadsto) = \sum_{(\x_i,B_i)\in S}\glo(h,\leadsto,\x_i,B_i).\]

Similar to previous sections, we now want to define a loss class for $\H \times \G$. We define $g(h,\leadsto)$ to be the set of all pairs $(\x,B)\in \X \times 2^{\X}$ on which $\glo(h,\leadsto,\x,B) =1$. Then the \emph{graph loss class} of $\H \times \G$ is defined as
\[(\H \times \G)_{\glo} = \{g(h,\leadsto) : h\in \H \text{ and } \leadsto \in \G\}.\]
We will now show that if the VC-dimension of the loss class $(\H \times \G)_{\glo}$ is finite, we can indeed learn $\mathcal{G}$ with respect to $\glo$. For some examples and more discussion on the VC-dimension with respect to the loss class $(\H \times \G)_{\glo}$, we refer the reader to the appendix.
\begin{lemma}\label{lem:graphlearning}
Let $\vc((\H \times \G)_{\glo}) =d$. Then there is $n_{\mathrm{graph}}: (0,1)^2\mapsto \mathbb{N}$, such that for any marginal distribution $P_\X$ and any manipulation graph $\to$ for a sample $S =\{(\x_1,B_{\to}(\x_1)),\dots,(\x_n,B_{\to}(\x_n)) \} $ of size $n\geq n(\epsilon,\delta)$, we have with probability at least $1-\delta$ over the sample generation $S_\X = (\x_1,\dots, \x_n)\sim \marginal^n$ for any $h\in \H$ and any $\leadsto \in \G$
\[  | \gLo{(P_\X, \to)}(h,\leadsto) - \gLo{S}(h,\leadsto)| < \epsilon .\]
Furthermore, $n_{\mathrm{graph}}(\epsilon,\delta)\in O(\frac{d+ \log \frac{1}{\delta}}{\epsilon^2})$.
\end{lemma}
We note that the above lemma is agnostic in the sense that it did not require $\to \in \G$.
We will now introduce an empirical version of the $\H$-$\P_\X$-distance. This will allow us to state the main theorem of this section and show that it is indeed possible to learn $\leadsto \in \G$ with low $\dHPX(\to,\leadsto)$ if $\vc((\H \times \G)_{\glo})$ is finite.
\begin{definition}
Given a sample $S_\X = \{(\x_1,\dots,\x_n)\}$ of domain elements $\x_i$ and two manipulation graphs $\to$ and $\leadsto$ we can define the empirical $\H$-$S_\X $-distance as
\[ d_{\H,S_\X}(\to, \leadsto) = \sup_{h\in \H} \sum_{\x_i\in S_{\X}} \glo(h,\leadsto,\x_i,B_{\to}(\x_i))  .\]
\end{definition}

\begin{theorem}\label{thm:dHPXlearning}
Let $\vc((\H \times \G)_{\glo}) =d$. Then there is $n_{\mathrm{dist}}: (0,1)^2\mapsto \mathbb{N}$, such that for every marginal distribution $P_\X$ and every manipulation graph $\to$ for a sample $S =\{(\x_1,B_{\to}(\x_1)),\dots,(\x_n,B_{\to}(\x_n)) \} $ of size $n\geq n(\epsilon,\delta)$, we have with probability at least $1-\delta$ over the sample generation $S_\X = (\x_1,\dots, \x_n)\sim \marginal^n$ for any  $\leadsto \in \G$
\[   d_{\H,P_{\X}}(\to,\leadsto) <  d_{\H,S_{\X}}(\to,\leadsto) + \epsilon .\]
Furthermore, $n_{\mathrm{dist}}(\epsilon,\delta)\in O(\frac{d+ \log \frac{1}{\delta}}{\epsilon^2})$.
\end{theorem}

Combining Theorem~\ref{thm:dHPXlearning} and Theorem~\ref{thm:approximategraphbound} we can thus conclude that it is indeed possible to learn $\leadsto \in \G$ such that using $\sloprime$ as a surrogate loss function guarantees a good approximation on the true strategic loss $\slo$. 

\section{Conclusion}
In this paper, we introduced a new strategic loss, which incentivizes correct classification under strategic manipulations. We also incorporate the idea of social burden into our notion of loss. We differentiated this loss from previous formal frameworks designed to mitigate strategic manipulation. In particular, we showed that optimizing for our strategic loss can yield satisfactory classification rules, even if there is no incentive-compatible hypothesis in the class that performs well on the classification task at hand. In addition, the loss formulation yields desirable effects in terms of sample complexity. Our work opens various avenues for further investigations and we hope it will inspire follow-up studies on the connections between a hypothesis class and the underlying manipulation graphs, effects of these connections, as well as learnability of the manipulation graph.

\section{Acknowledgements}
Ruth Urner is also a faculty affiliate member of Toronto's Vector Institute. Her research is supported by an NSERC Discovery grant. Tosca Lechner is also a graduate student of the Vector Institute and was supported by a Vector Research Grant.

\bibliography{strategic_refs}

\newpage
 \appendix
 \section{Statistical Learning Theory basics}
In this section, we detail the standard setup of statistical learning theory for classification which we employ in our work.
\subsection{Learning theoretic setup and definitions}
We let  $\X\subseteq\reals^d$ denote the domain and $\Y$ (mostly $\Y=\{0,1\}$) a (binary) label space.
We model the data generating process as a distribution $P$ over $\X\times \Y$ and let $P_\X$ denote the marginal of $P$ over $\X$.
We use the notation $(\x,y)\sim P$ to indicate that $(\x,y)$ is a sample from distribution $P$ and $S\sim P^n$ to indicate that set $S$ is a sequence (for example a training or test data set) of $n$ \iid samples from $P$.
Further, we use notation $\eta_P(\x) = \Pr_{(\x,y)\sim P}[y = 1 \mid \x]$ to denote the \emph{regression} or \emph{conditional labeling function} of $P$.
We say that the distribution has \emph{deterministic labels} if $\eta_P(\x) \in \{0,1\}$ for all $\x\in\X$.

A \emph{classifier} or \emph{hypothesis} is a function $h:\X\to\Y$.
A classifier $h$ can naturally be viewed a subset of $\X\times \Y$, namely $h = \{(\x,y)\in X\times Y ~\mid~ \x\in X,~y = h(\x)\}$.
We let $\F$ denote the set of all Borel measurable functions from $\X$ to $\Y$ (or all functions in case of a countable domain). 
A \emph{hypothesis class} is a subset of $\F$, often denoted by $\H\subseteq \F$.

The quality of prediction of a hypothesis on an input/output pair $(x,y)$ is measured by a \emph{loss function} $\lo:(\F\times \X \times \Y) \to \reals$.
For classification problems, the quality of prediction is typically measured with the \emph{binary} or \emph{classification loss}
\[
\blo(h, x, y) = \indct{h(x) \neq y}, 
\]
where $\indct{\alpha}$ denotes the indicator function for predicate $\alpha$.

We denote the \emph{expected loss} (or \emph{true loss}) of a hypothesis $h$ with respect to the distribution $P$ and loss function $\lo$ by 
\[
\Lo{P} (h) = \Ex_{(x,y)\sim P} [\lo(h , x, y)].
\]
In particular, we will denote the true binary loss of a classifier $h$ by $\bLo{P}(h)$.
The quality of an output classifier is typically compared with the best possible (binary) loss achievable on distribution $P$ with hypotheses from a fixed class $\H$. This quantity is referred to as the \emph{approximation error of $\H$ with respect to $P$} and denoted by 
\[
\bopt{P}(\H) = \inf_{h\in \H} \bLo{P}(h).
\]

The \emph{empirical loss} of a hypothesis $h$ with respect to loss function $\lo$ and a sample $S = ((x_1, y_1), \ldots, (x_n, y_n))$ is defined as 
\[
\Lo{S}(h) = \frac{1}{n}\sum_{i=1}^n \lo(h, x_i, y_i).
\]

A \emph{learner} $\A$ is a function that takes in a finite sequence of labeled instances $S = ((x_1, y_1), \ldots, (x_n, y_n))$ and outputs a hypothesis $h = \A(S)$.
The following is a standard notion of (PAC-)learnability of a hypothesis class from finite samples
\cite{vapnikcherv71,Valiant84,blumer1989learnability,shalev2014understanding}.

\begin{definition}[PAC learnability]\label{def:learnability}
 We say that a learner $\A$ is \emph{PAC learns} (or simply \emph{learns}) hypothesis class $\H$ with respect to a set of distributions $\P$ and loss function $\lo$ if, for every $\epsilon,\delta >0$ there is a sample-size $n(\epsilon, \delta)$ such that, for all $n \geq n(\epsilon, \delta)$, we have
 $$\Pr_{S\sim P^n}\left[ \Lo{P}(\A(S)) \leq \opt{P}(\H) + \epsilon \right] \geq 1-\delta.$$
 We say that $\A$ is \emph{agnostically learns $\H$}, if the above holds with respect to the class of all data-generating distributions (subject to some mild, standard measurability conditions). We say that $\A$ learns $\H$ \emph{in the realizable case} if the above holds with respect to the class $\P$ of distributions for which there exists a classifier $h^*\in \H$ with $\Lo{P}(h^*) =0$.
\end{definition}

The smallest function $n:[0,1]^2 \to \naturals$ for which there exists a learner $\A$ that learns class $\H$ in the above sense is called the \emph{sample complexity of $\H$}.
\begin{definition}[Proper versus improper learning]\label{def:proper}
If a learner $\A$ always outputs a function $\A(S)\in\H$ from the hypothesis class $\H$, we call $\A$ a \emph{proper learner for $\H$} (and otherwise we call $\A$ and \emph{improper learner for $\H$}). If Definition \ref{def:learnability} for learnability holds with a proper learner for $\H$, we call the class $\H$ \emph{proper (PAC) learnable}.
\end{definition}
It is well known that a binary hypothesis class $\H$ is learnable (with a proper learner; both agnostically and in the realizable case) in the above sense if and only if $\H$ has finite VC-dimension $\vc(\H)$ (see Definition \ref{def:vc} in the main part of the paper, and more details on the background of this in Section on loss classes); and that the sample complexity is $\tilde{\theta}\left(\frac{\vc(\H) \log(1/\delta)}{\epsilon}\right)$ in the realizable case and ${\theta}\left(\frac{\vc(\H) \log(1/\delta)}{\epsilon^2}\right)$ in the agnostic case \cite{shalev2014understanding}.

\subsection{The role of VC-dimension of loss classes for learnability}
Standard VC-theory tells us that, for the binary classification loss, the sample complexity of learning a hypothesis class is determined by the VC-dimension of the class (and thus the VC-dimension of the loss class since these are identical for the binary loss). Any learner that acts according to the  ERM (Empirical Risk Minimization) principle is a successful learner with respect to the binary loss for classes of bounded VC-dimension $d$. We here briefly recap the underpinnings of this result. For the realizable case, a sample $S\sim P^n$ of size at least $\tilde{O}(d \log(1/\delta)/\epsilon)$ is an $\epsilon$-net for the loss class (with probability at least $1-\delta$) \cite{HausslerW87}, thus, for every hypothesis that has loss at least $\epsilon$, there would be a sample point indicating that. Choosing a hypothesis of zero empirical loss, then guarantees true loss bounded by $\epsilon$. In the agnostic case, a sample of size ${O}(d \log(1/\delta)/\epsilon^2)$ is an $\epsilon$-approximation for the loss class, that is, we have $|\Lo{P}(h) - \Lo{S}(h)| \leq \epsilon$ for every $h\in \H$ with high probability, which in turn implies
that any ERM learner is a successful agnostic PAC learner. For the binary loss there are complementing lower bounds for the sample complexity of learning VC-classes.
However, unlike for the binary loss, the VC-dimension of the strategic loss class does not imply a lower bound on the sample complexity (as we see in Theorem~\ref{thm:nolowerbound}).

\section{Example for bounded VC dimension of strategic component}\label{s:examples}
To illustrate the conditions in Theorem \ref{thm:boundedvc},  we here provide a few natural examples, where the VC-dimension of the strategic components can be bounded:
\begin{example}
\begin{enumerate}
    \item  $\M = (\X, \to)$ being complete, implies $\vc(\H_{\sclo}) = \vc(\H)$.
    \item If $\M$ corresponds to a partial order over $\X$ (that is, in particular $\M$ is acyclic) and the class $\H$ is a subset of complements of initial segments in $\M$ (that is if, for some $h\in\H$ and $\x\in\X$ we have $h(\x) = 0$, then we also have $h(\x')= 0$ for all $\x'$ that precede $\x$ in the partial order) , then the VC-dimension of $\H_{\sclo}$ is bounded by the size of a largest antichain in $\M$.
    \item If $\X = \reals^d$, $\H$ consists of linear classifiers and the plausible manipulations $\x \to \x'$ are determined by $\|\x - \x'\|_p \leq r$ two points being close in some standard norm, then the VC-dimension of $\H_{\sclo}$ is bounded by $2d +2$.
    \item If $\X = \reals^d$, $\H$ consists of linear classifiers and we have plausible manipulations $\x \to \x'$ if and only if $\x$ and $\x'$ differ on only one coordinate. Then the VC-dimension of the resulting class of strategic components is $d+1$ (since for every $h$, for every point $\x$ such that $h(x)=0$, there is a $\x'$ such that $\x\to\x'$ is an edge and $h(\x') =1$. Thus, for every half-space $h$, sets of zeros of $h$ that are connected to a $1$ is exactly $\{\x: h(\x)=0\}$, and therefore $\vc(\H_{\sclo}) = \vc(\H)$.
\end{enumerate}
\end{example}

\section{The VC dimension of the graph loss class}
In Section~\ref{sec:graphlearning} we analyse the learnability of an approximate manipulation graph from a predefined set of manipulation graphs $\G$ in $\H$-$\marginal$-distance in terms of the VC dimension of the graph loss class $(\H \times \G )_{\glo} $. We now want to give some examples for when this VC dimension is indeed finite. 
\begin{example}
\begin{enumerate}
    \item If for every $h\in \H$ and every $\to \in \G$, we have $VC((\{h\}\times \G)_{glo})\leq d_1 $ and $VC(( \H \times \{\to\})_{\glo})\leq d_2 $ then $VC((\H \times \G)_{\glo}) \leq d_1 d_2$.
    \item Let $\Hcal_{lin}=\{h: \mathbb{R}^d \rightarrow \{0,1\}: \exists w\in \mathbb{R}^d: h(x) = 1 \text{ if and only if } x^Tw \geq 0\}$ be the class of linear separators and $\mathcal{G}_{lin}= \{\to: \exists w\in \mathbb{R}^d:  f_{\to}(x) = B_{x^tw}(x)\}$ be the graph class consistent of graphs such that all their neighborhood sets are balls whose radius may depend linearly on the feature vector. Then the VC-dimension of $\Hcal_{lin}\times \mathcal{G}_{lin}$ is finite.
\end{enumerate}
\end{example}

 \section{Proofs}
 \begin{repobservation}{obs:vcgrows} 
For any $d\in\naturals\cup\{\infty\}$ there exists a class $\H$ and a manipulation graph $\M = (\X, \to)$ with $\vc(\H) =1 $ and $\vc(\H_\slo) \geq d$.
\end{repobservation}

\begin{proofof}{Observation~\ref{obs:vcgrows}}
Let $\X$ be some infinite domain. We treat the case $d\in\naturals$ first.
We consider a set a set $\U = \{\x_1, \x_2, \ldots, \x_d\}$ of $d$ domain points and a disjoint set $\V = \{\z_1, \z_2, \ldots, \z_{2^d}\}$ of $2^d$ domain points. We associate each subset of $\U$ with exactly one point in  $\V$. We design a manipulation graph $\M$ by adding an edge $\x_i \to \z_j$ if and only if $\x_i$ is a member of the subset associated with $\z_j$, and include no other edges. Now we consider the class $\H$ of singletons over $\V$, that is $\H = \{h_1, h_2, \ldots, h_{2^d}\}$ consists of $2^d$ functions and we have $h_j(\x) = \indct{\x = \z_j}$. That is $h_j$ assigns label $0$ to every point in the domain except for $\z_j$. Then we have $\vc(\H) = 1$. However the loss class $\H_{\slo}$ shatters the set  $\{(\x_1, 0), (\x_2,0) \ldots, (\x_d, 0)\}$ (and also the set $\{(\x_1, 1), (\x_2,1) \ldots, (\x_d, 1)\}$). Thus, $\vc(\H_{\slo})\geq d$.
For the case $d=\infty$, we will use the same construction with the set $\U=\naturals$ and $\V$ being an uncountable set, again indexed by the power-set of $\U$.
\end{proofof}

\begin{repobservation}{obs:VCinequ}
For any hypothesis class $\H$ and any manipulation graph $\M = (\X, \to)$, we have $\vc(\H) \leq \vc(\H_\slo)$.
\end{repobservation}

\begin{proofof}{Observation~\ref{obs:VCinequ}}
Let $\{(\x_1, y_1), (\x_2, y_2) \ldots, (\x_d, y_d)\}$ be a set of points shattered by the hypothesis class $\H$.
Since all elements of $\H$ are functions, we can assume $y_i = 1$ for all $i$. 
The same set of domain points  $\{(\x_1, 1), (\x_2, 1) \ldots, (\x_d, 1)\}$ with all labels set to $1$  is then shattered by the loss class $\H_\slo$. To see this, note that for a function $h\in \H$ and point $\x$, if $h(\x) = 1$, then $(\x,1) \notin h_\slo$ is not in the loss set. Thus, if the class $\H$ can shatter those points, the loss class will shatter them as well.
\end{proofof}

\begin{reptheorem}{thm:nolowerbound}
For every $d\in\naturals\cup\{\infty\}$, there exists a hypothesis class $\H$ with $\vc(\H_\slo) = d$ that is learnable with sample complexity $O(\log(1/\delta)/\epsilon)$ in the realizable case.
\end{reptheorem}

\begin{proofof}{Theorem~\ref{thm:nolowerbound}}
We consider the class of singletons from the proof of Observation \ref{obs:vcgrows}. Note that, if the class is realizable with respect to the strategic loss, then there exists at most one $\z_j$ such that this point $(\z_j, 1)$ with label $1$ has a positive weight under this distribution.
Further, the distribution can not assign any weight to points $(\x_i, 1)$, and it can only assign a positive weight to points $(\x_i , 0)$ if $\x_i$ is not a member of the subset corresponding to $\z_j$. Further, if the point $(\z_j, 1)$ is a member of the training sample, the learner can output hypothesis $h_j$ and this hypothesis then has true strategic loss $0$. If the training sample does not contain any point with label $1$, the learner can output a hypothesis $h_0$ that assigns label $0$ to all points. In that case, we get $\sLo{P}(h_0) = P(\z_j, 1)$. If the probability $P(\z_j, 1)$ was larger than $\epsilon$, then, by the standard argument, a sample of the stated size being an $\epsilon$-net for singletons would have contained the point (with high probability at least $1-\delta$).
\end{proofof}

\begin{reptheorem}{thm:adaptationofmontasser}{\bf (Adaptation of Theorem 1 and Theorem 4 in \cite{MontasserHS19})}\\
There exists a hypothesis class $\H$ with $\vc(\H) = 1$ that is not learnable with respect to the strategic loss by any proper learner $\A$ for $\H$ even in the realizable case. On the other hand, every class $\H$ of finite  VC-dimension is learnable (by some improper learner).
\end{reptheorem}

\begin{proofof}{Theorem~\ref{thm:adaptationofmontasser}}
The construction to prove the first claim can be taken exactly as is in the proof of Theorem 1 by Montasser et al.\cite{MontasserHS19} by setting the label $+1$ there to 0 and the label $-1$ there to $1$. The positive result on general learnability with an improper learner is derived by Montasser et al.\cite{MontasserHS19} by means of adapting a general compression scheme for VC-classes developed by Moran and Yehudayoff \cite{moran2016sample}. We can use the same adaptation to show learnability with respect to the strategic loss with the following modification: instead of just compressing a data sample $S$, the compression scheme for the adversarial loss first creates an inflated sample $S_U$ which adds the perturbation sets (and then discretizing these using Sauer's lemma to obtain a new finite data set). For the strategic loss, we would add the label conditioned perturbation sets (as described above), that is we inflate the sample by adding points $(\x', y)$, for $(\x_i, y)$ being the data point of smallest index with $\x_i\to \x'$. The discretization step and remaining compression and decompression technique can be used identically as in the adversarial loss case.
\end{proofof}

\begin{reptheorem}{thm:approximategraphbound}
Let $\mathcal{H}$ be any hypothesis class and $\to,\leadsto$ two manipulation graphs. Then for any distribution $P$ over $\X\times\Y$ and any $h\in \mathcal{H}$ we have
\begin{align*}
   \sLo{P}(h)  &\leq \bLo{P}(h) + \scLoprime{P}(h) + \dHPX(\to, \leadsto) \\
   &\leq  2 \sLoprime{P}(h) + \dHPX(\to, \leadsto) . 
\end{align*}
Furthermore, by rearranging the result, we get
\[ \frac{1}{2} \sLoprime{P}(h) - \dHPX(\to, \leadsto) \leq \sLo{P}(h). \]
\end{reptheorem}

\begin{proofof}{Theorem~\ref{thm:approximategraphbound}}
\begin{align*}
 \sLo{P}(h)  & =  \bLo{P}(h) + \scLo{P}(h)\\
 &- \mathbb{E}_{(x,y)\sim P}[h(x) =0, y=1, \exists x': x\to x': h(x') =1] \\
   &\leq \bLo{P}(h) + \scLoprime{P}(h) + |\scLo{P}(h) -\scLoprime{P}(h) |\\
   &\leq \bLo{P}(h) + \scLoprime{P}(h) + \dHPX(\to,\leadsto)\\
   & \leq 2 \sLoprime{P}(h) + \dHPX(\to, \leadsto)
\end{align*}
By exchanging $\to$ and $\leadsto$ and rearranging we get the second inequality.
\end{proofof}

\begin{replemma}{lem:graphlearning}
Let $\vc((\H \times \G)_{\glo}) =d$. Then there is $n_{\mathrm{graph}}: (0,1)^2\mapsto \mathbb{N}$, such that for any marginal distribution $P_\X$ and any manipulation graph $\to$ for a sample $S =\{(x_1,B_{\to}(x_1)),\dots,(x_n,B_{\to}(x_n)) \} $ of size $n\geq n(\epsilon,\delta)$, we have with probability at least $1-\delta$ over the sample generation $S_\X = (x_1,\dots, x_n)\sim P_X^n$ for any $h\in \H$ and any $\leadsto \in \G$
\[  | \gLo{(P_\X, \to)}(h,\leadsto) - \gLo{S}(h,\leadsto)| < \epsilon .\]
Furthermore, $n_{\mathrm{graph}}(\epsilon,\delta)\in O(\frac{d+ \log \frac{1}{\delta}}{\epsilon^2})$.
\end{replemma}

\begin{proofof}{Lemma~\ref{lem:graphlearning}}
The result follows directly from the finite VC-dimension of $\vc((\H \times \G)_{\glo})$ and the fact that a finite VC-dimension of a class yields uniform convergence of the class (Theorem 6.7 and Theorem 6.8 from \cite{shalev2014understanding}).
\end{proofof}

\begin{reptheorem}{thm:dHPXlearning}
Let $\vc((\H \times \G)_{\glo}) =d$. Then there is $n_{\mathrm{dist}}: (0,1)^2\mapsto \mathbb{N}$, such that for any marginal distribution $P_\X$ and any manipulation graph $\to$ for a sample $S =\{(x_1,B_{\to}(x_1)),\dots,(x_n,B_{\to}(x_n)) \} $ of size $n\geq n(\epsilon,\delta)$, we have with probability at least $1-\delta$ over the sample generation $S_\X = (x_1,\dots, x_n)\sim P_X^n$ for any  $\leadsto \in \G$
\[   d_{\H,P_{X}}(\to,\leadsto) <  d_{\H,S_{X}}(\to,\leadsto) + \epsilon .\]
Furthermore, $n_{\mathrm{dist}}(\epsilon,\delta)\in O(\frac{d+ \log \frac{1}{\delta}}{\epsilon^2})$.
\end{reptheorem}

\begin{proofof}{Theorem~\ref{thm:dHPXlearning}}
Let $\leadsto \in \G$ and $S_\X \sim P_\X^n$. Then using Lemma~\ref{lem:graphlearning} we get with probability $1-\delta$
\begin{align*}
   & | d_{\H,P_{X}}(\to,\leadsto) - d_{\H,S_{X}}(\to,\leadsto)| \\
   &= | \sup_{h\in \mathcal{H}} \mathbb{E}_{x\sim P_\X}[|\sclo(h,x) - \scloprime(h,x) |]\\
   &-  \sup_{h\in \H} \sum_{x_i\in S_{\X}} \glo(h,\leadsto,x_i,B_{\to}(x_i))|\\
    &\leq \sup_{h\in \H } |  \mathbb{E}_{x\sim P_\X}[|\sclo(h,x) - \scloprime(h,x) |]\\
    &-  \sum_{x_i\in S_{\X}}  \glo(h,\leadsto,x_i,B_{\to}(x_i))| \\
    &\leq \sup_{h\in \H } |  \gLo{(P_\X,\to)}(h,\leadsto) -\gLo{S}(h,\leadsto)| < \epsilon .
\end{align*}
\end{proofof}

\end{document}